\documentclass[letterpaper]{article}

\usepackage{natbib,alifeconf}  
\usepackage{url,hyperref}
\usepackage{booktabs}

\usepackage{amsmath}
\usepackage{cleveref}
\usepackage{float}
\usepackage{microtype}
 
\title{Toward Artificial Open-Ended Evolution within Lenia using Quality-Diversity}

\author{
    Maxence Faldor, \and
    Antoine Cully\\
    \mbox{}\\
    Imperial College London, United Kingdom\\
    m.faldor22@imperial.ac.uk\\
    \href{https://leniabreeder.github.io}{leniabreeder.github.io}
} 

\begin{document}

\newcommand{\ours}{Leniabreeder}

\newcommand*\diff{\mathop{}\!\mathrm{d}}

\newcommand{\lattice}{\mathcal{L}}
\newcommand{\world}{\mathbf{A}}
\newcommand{\kernel}{\mathbf{K}}
\newcommand{\potential}{\mathbf{U}}
\newcommand{\growth}{\mathbf{G}}
\newcommand{\site}{x}

\newcommand{\me}{MAP-Elites}
\newcommand{\aurora}{AURORA}
\newcommand{\solution}{\mathbf{x}}
\newcommand{\latent}{\mathbf{z}}
\newcommand{\params}{\theta}

\maketitle

\begin{abstract}
From the formation of snowflakes to the evolution of diverse life forms, emergence is ubiquitous in our universe.
In the quest to understand how complexity can arise from simple rules, abstract computational models, such as cellular automata, have been developed to study self-organization.
However, the discovery of self-organizing patterns in artificial systems is challenging and has largely relied on manual or semi-automatic search in the past.
In this paper, we show that Quality-Diversity, a family of Evolutionary Algorithms, is an effective framework for the automatic discovery of \emph{diverse} self-organizing patterns in complex systems.
Quality-Diversity algorithms aim to evolve a large population of diverse individuals, each adapted to its ecological niche.
Combined with Lenia, a family of continuous cellular automata, we demonstrate that our method is able to evolve a diverse population of lifelike self-organizing autonomous patterns.
Our framework, called \ours{}, can leverage both manually defined diversity criteria to guide the search toward interesting areas, as well as unsupervised measures of diversity to broaden the scope of discoverable patterns.
We demonstrate both qualitatively and quantitatively that \ours{} offers a powerful solution for discovering self-organizing patterns.
The effectiveness of unsupervised Quality-Diversity methods combined with the rich landscape of Lenia exhibits a sustained generation of diversity and complexity characteristic of biological evolution.
We provide empirical evidence that suggests unbounded diversity and argue that \ours{} is a step toward replicating open-ended evolution in silico.
\end{abstract}

\section{Introduction}
Over four billion years, evolution on Earth has showcased a captivating trend of continuous innovation and increasing biological complexity. This phenomenon has intrigued scientists and sparked the fundamental question of how inanimate matter can spontaneously organize into a diversity of life forms. To understand how complex wholes can emerge from simple parts, researchers have turned to computational models.

Pioneered by John von Neumann and others, particularly through groundbreaking work on self-replicating machines~\citep{automata}, Cellular Automata (CA) have become a fundamental framework for studying emergence and complexity. CA are capable of generating complex patterns that emerge solely from the local interactions of their components, following simple, deterministic rules. Conway's Game of Life~\citep{cgol} is a prominent example among CA. Despite its underlying simplicity --- defined by a set of four basic rules governing the birth, survival, and death of cells on a grid --- Conway's Game of Life has given rise to a surprisingly vast array of self-organizing structures (e.g., stable forms, oscillators, spaceships, etc.). Later on, it was proved to be Turing complete, a property meaning it can simulate any Turing machine.

Continuous CA, such as Lenia~\citep{chan2019lenia} and SmoothLife~\citep{rafler_GeneralizationConwayGame_2011}, marked a significant advancement by bridging the gap between the discrete nature of Conway's Game of Life and the continuous dynamics characteristic of the real world. Interactive evolutionary computation methods revealed that Lenia can support a diversity of lifelike, self-organizing autonomous patterns. It led to the identification and classification of hundreds of artificial species, uncovering emergent behaviors such as locomotion, differentiation, reproduction, and emission~\citep{chan2019lenia,chan_LeniaExpandedUniverse_2020}. Consequently, Lenia stands as a fertile ground for exploring the underlying mechanisms of artificial evolution within a controlled computational environment and serves as an ideal testbed for examining the emergence of diverse artificial life. However, self-organizing patterns have mostly been discovered through manual or semi-automatic search algorithms~\citep{chan2019lenia}, limiting our ability to fully explore this vast potential.

Mirroring biological evolution, Quality-Diversity~\citep{pugh_QualityDiversityNew_2016,cully_QualityDiversityOptimization_2018} is a family of Evolutionary Algorithms that aims to discover a diverse population of individuals, each adapted to its ecological niche. In contrast with traditional optimization methods, the goal of Quality-Diversity algorithms is to find a large collection of different, high-performing solutions. Consequently, these methods hold the promise to realize Lenia's full potential and illuminate an ecosystem of diverse artificial species. Objective-based optimization methods are prone to get stuck in local optima, whereas keeping a repertoire of diverse solutions can help to find stepping stones that lead to globally better solutions~\citep{mouret_IlluminatingSearchSpaces_2015,dcgme,faldor_SynergizingQualityDiversityDescriptorConditioned_2023a}, mimicking evolution in nature.

In this work, we show that Quality-Diversity algorithms are an effective solution to the problem of automatic discovery of \emph{diverse} self-organizing patterns in high-dimensional complex systems~\citep{reinke_IntrinsicallyMotivatedDiscovery_2020,etcheverry_HierarchicallyOrganizedLatent_2021}. In particular, we demonstrate that Quality-Diversity has the capacity to unleash the untapped potential of Lenia's rich landscape. To that end, we leverage both supervised and unsupervised Quality-Diversity methods. The supervised approach, \me{}~\citep{mouret_IlluminatingSearchSpaces_2015}, utilizes manually defined diversity metrics to guide the exploration toward specific characteristics of interest, facilitating the identification of patterns with unique properties such as color or motion. However, the necessity to manually specify diversity criteria inherently restricts the breadth of discoverable self-organizing patterns. To address this limitation, we employ an unsupervised approach, \aurora{}~\citep{grillotti_UnsupervisedBehaviorDiscovery_2022}, that automatically learns a measure of diversity, significantly broadening the scope of discoverable patterns without the need for predefined diversity criteria.

However, this unsupervised method comes with its own set of challenges. Indeed, the Lenia search space is vast, and this artificial evolution process will likely lead to diverse patterns that explode or evaporate quickly. Although some of these individuals present intriguing similarities to Turing patterns~\citep{turing}, in this paper, we focus the search toward \emph{localized} and \emph{autonomous} self-organizing  patterns, called \emph{solitons}~\citep{chan2019lenia}. To that end, we introduce a set of both manually defined and unsupervised fitness functions, that capture basic characteristics of life, such as agency or homeostasis~\citep{lyfe}. These fitness functions encode simple heuristics that guide the search toward meaningful expressions of artificial life.


We introduce \ours{}, a framework designed to automate the discovery of \emph{diverse} self-organizing patterns in complex systems. Our contributions are as follows:
\begin{itemize}
    \item We show that Quality-Diversity is an effective approach for the automatic discovery of diverse artificial species within Lenia. Those methods are generally applicable to other artificial life systems.
    \item We introduce a set of manually defined and unsupervised fitness and descriptor functions, tailored to guide the search toward meaningful expressions of artificial life.
    \item We provide evidence that our method demonstrates some characteristics of artificial open-ended evolution, exhibiting a sustained generation of diversity and mirroring the continuous innovation observed in nature.
\end{itemize}
We report quantitative and qualitative results, underscoring the potential of our framework to unlock new frontiers in artificial life research. Through the convergence of Lenia and Quality-Diversity, we explore open-ended evolution within computational systems.

\section{Background}
\subsection{Lenia}
Lenia is a cellular automaton that generalizes Conway's Game of Life to continuous space-time-state, generalized local rule as well as higher dimensions, multiple kernels, and multiple channels~\citep{chan_LeniaExpandedUniverse_2020}. Interactive evolutionary computation methods have revealed that Lenia supports a diversity of lifelike self-organizing autonomous patterns~\citep{chan2019lenia}, making it a fertile ground for the study of artificial open-ended evolution~\citep{chan_LargeScaleSimulationsOpenEnded_2023}.

In Lenia, the world starts in an initial configuration $\world^0$, defined as a $d$-dimensional lattice with $c$ channels of real values between $0$ and $1$. In this work, we use the generalized rule with multiple kernels $\kernel_k$ and growth mappings $G_k$~\citep{chan_LeniaExpandedUniverse_2020}. The update is calculated as an average of the results for each kernel, weighted by factors $h_k/h$. Therefore, the state of the world is updated according to the formula:
$$\world_j^{t+\Delta t} = [\world_j^t + \Delta t \sum_{i,k} \frac{h_k}{h} G_k(\kernel_k * \world_i^t)]_0^1$$

Each kernel is characterized by a relative radius $r_k R$, a parameter $\beta_k \in [0, 1]^B$ and a growth mapping with parameters $\mu_k$ and $\sigma_k$ following the descriptions by~\citet{chan_LeniaExpandedUniverse_2020,chan2019lenia}
. In this paper, a growth mapping is a function $G_k: [0, 1] \rightarrow [-1, 1]$ such that $G_k(u) = 2 \exp(-\frac{1}{2}(\frac{u-\mu_k}{\sigma_k})^2) - 1$ and a kernel $\kernel_k$ is constructed by combining an exponential kernel core and a kernel shell with parameter $\beta_k$, as defined by~\citet{chan2019lenia}. To summarize, each kernel is defined by a set of parameters $(r_k R, \beta_k, \mu_k, \sigma_k, h_k)$.

\subsection{Quality-Diversity}
Evolution in nature has the remarkable capacity to produce a rich diversity of species, each exquisitely adapted to its local environmental niche. Inspired by this idea, Quality-Diversity approaches, such as novelty search with local competition~\citep{lehman_EvolvingDiversityVirtual_2011} or \me{}~\citep{mouret_IlluminatingSearchSpaces_2015}, are a family of evolutionary algorithms that aim to return a collection of different niches, as well as the best individual living in each niche~\citep{pugh_QualityDiversityNew_2016}.
In contrast with traditional evolutionary algorithms that focus solely on finding the optimal solution, Quality-Diversity methods generate large populations of simultaneously high-fitness and different individuals~\citep{cully_QualityDiversityOptimization_2018}.

In addition to the fitness $F(\solution)$ that determines the quality of a solution $\solution$ in the search space $\mathcal{X}$, Quality-Diversity optimization also necessitates the descriptor $D(\solution)$, that is generally manually defined and characterizes the solution $\solution$ for the type of diversity desired. The descriptor space $\mathcal{D} = D(\mathcal{X})$ together with the Euclidean distance define a metric space that enables the computation of distances between individuals or to measure the novelty of a new solution. The objective of QD algorithms is to find the highest fitness solution at each point of the descriptor space.

\subsubsection{\me}
\me{}~\citep{mouret_IlluminatingSearchSpaces_2015} is a simple but efficient Quality-Diversity method. The algorithm discretizes the descriptor space into a multi-dimensional grid of cells and searches for the best solution in each cell. \me{} starts by initializing the grid with random solutions. Then, the algorithm iteratively executes the following steps until a predefined budget of evaluation is reached: (1) a batch of parent solutions is uniformly selected from the grid, (2) a batch of offspring solutions is generated from the parents through a variation operator, (3) for each offspring solution, both its fitness and descriptor are evaluated, and (4) offspring solutions are added to the grid. A solution is added to its corresponding cell in the grid if and only if the cell is empty or the solution has a higher fitness than the current solution occupying that cell, in which case the current solution is replaced by the new one.

\subsubsection{\aurora}
\aurora{}~\citep{cully_AutonomousSkillDiscovery_2019,grillotti_UnsupervisedBehaviorDiscovery_2022} is a Quality-Diversity algorithm that circumvents the need for manual definitions of diversity. It leverages unsupervised learning to automatically define a descriptor function through parameters $\params$. The algorithm follows a standard QD loop. During the evaluation step, the descriptors are determined using the current descriptor function, denoted as $D_\params$. In turn, this descriptor function is continuously trained via unsupervised learning, on the data generated during evaluation, giving new parameters $\params^\prime$. The new descriptor function $D_{\params^\prime}$ redefines the niches within the search space, not only influencing local competition within the current population but also shaping subsequent offspring evaluation and addition.
This \emph{dynamic} interaction between the individuals and their niches propels a cycle of discovery, where each individual adapts to its niche but also drives the realignment of niche boundaries. \aurora{} mirrors the dynamic interplay observed in ecosystems, where species not only adapt to environmental changes but also shape the structure and boundaries of niches~\citep{jones_JonesCGLawton_1994,wright_EcosystemEngineerBeaver_2002}.

\subsection{Open-Ended Evolution}
Open-ended evolution research seeks to understand the mechanisms and conditions that enable the perpetual emergence of novelty, characteristic of biological evolution~\citep{packard_OverviewOpenEndedEvolution_2019}. This pursuit is one of the greatest challenges in evolutionary biology and artificial life research. Exploring open-ended evolution goes beyond understanding biology or generating captivating simulations. It raises fundamental questions about the nature of creativity, the emergence of complexity, and how innovative solutions can arise in artificial systems~\citep{soros_OpenendednessLastGrand_2017}.

Despite significant efforts, achieving genuine open-ended evolution in silico has proven challenging. A key obstacle is the development of formal and objective definitions, despite some advancements in this area~\citep{packard_OverviewOpenEndedEvolution_2019,adams_FormalDefinitionsUnbounded_2017,taylor2015requirements,maley,hintze_OpenEndednessSakeOpenEndedness_2019,soros_IdentifyingNecessaryConditions_2014,pattee_EvolvedOpenEndednessNot_2019}. Therefore, our work does not claim to achieve theoretical open-endedness. Instead, we offer it as a step toward understanding and replicating aspects of open-ended evolution.

Nature encompasses numerous niches, enabling different species to be successful independently of each other. For instance, the agility of cheetahs in hunting does not prevent orcas from thriving underwater. This remarkable range of species produced within a single run inspired Quality-Diversity algorithms to prioritize generating diverse solutions in pursuit of open-endedness.
However, \me{} does not allow the addition of new cells over time that did not exist in the original descriptor space, and as a result, cannot exhibit true open-ended evolution~\citep{mouret_IlluminatingSearchSpaces_2015}.
\aurora{} moves closer to open-ended evolution by dynamically adapting the criteria through which diversity is assessed. This method allows to continuously uncover new niches without being constrained by a predefined descriptor space, thereby avoiding premature convergence. As a result, \aurora{} can foster a more genuinely open-ended process of discovery, akin to the way biological evolution endlessly explores new forms of life and strategies for survival. Finally, this approach overcomes the limitations of manually defined diversity and complexity metrics that have historically hindered open-ended evolution research~\citep{hintze_OpenEndednessSakeOpenEndedness_2019}.

\section{Related Work}
\subsection{Automatic Discovery of Self-Organizing Patterns}
Cellular automata have been extensively studied to explore artificial self-organizing patterns, typically identified manually or assisted with computer simulation~\citep{automata,turing,wolfram}. While more refined methods have been employed to search for specific rules or patterns~\citep{sapin_ResearchCellularAutomaton_2003,mitchell_EvolvingCellularAutomata_2000}, our approach aims to illuminate a diversity of autonomous configurations. This focus on diversity aligns closely with methods like IMGEP, which have similarly been applied to discover a range of patterns within Lenia~\citep{etcheverry_HierarchicallyOrganizedLatent_2021,reinke_IntrinsicallyMotivatedDiscovery_2020}. Despite these similarities, our method diverges from IMGEP-based approaches in two key aspects. We adopt a methodology that is not goal-directed, making it inherently closer to the nature of biological evolution. Additionally, we compute fitness and descriptor based on multiple timesteps and not on the final state of the system, similar to~\citet{jain_CapturingEmergingComplexity_2024,cisneros_VisualizingComputationLargescale_2020,cisneros_EvolvingStructuresComplex_2019}.

Neural cellular automata have emerged as a powerful tool for studying regeneration and pattern formation~\citep{mordvintsev2020growing}. This approach, combining the principles of cellular automata with the adaptability of neural networks, has provided insights into the self-organizing capabilities of biological and artificial systems~\citep{mordvintsev2020growing,palm_VariationalNeuralCellular_2021}. Unlike our approach, which explores a wide range of self-organizing structures, these systems focus on growing specific target patterns.

\subsection{Artificial Open-Ended Evolution}
Numerous artificial life systems have demonstrated a potential for open-endedness~\citep{soros_IdentifyingNecessaryConditions_2014,standish_OpenEndedArtificialEvolution_2002,hintze_OpenEndednessSakeOpenEndedness_2019}. While most research in this area has focused on developing new algorithms, crafting novel artificial systems, or establishing necessary conditions for open-endedness, our work adopts a different approach. We show that existing Quality-Diversity algorithms, under appropriate conditions, inherently possess the capabilities to exhibit some open-ended evolutionary dynamics.

In the context of Lenia, recent research has investigated the potential for open-endedness. In particular, \citet{plantec_FlowLeniaOpenendedEvolution_2023} introduced a mass-conservative extension of Lenia, diverging from our method by modifying Lenia's rules. \citet{chan_LargeScaleSimulationsOpenEnded_2023} explored open-endedness through large-scale simulations, a different angle compared to our focus on the intrinsic dynamics of evolution without scaling the system's size.

\section{Methods}
We introduce \ours{}, a framework designed to automate the discovery of diverse patterns in complex systems. While we showcase its effectiveness in the context of Lenia, the methodology is generally applicable to other artificial life systems.
We formalize the discovery of diverse artificial species as an evolutionary algorithm, specifically a Quality-Diversity optimization problem. We employ two approaches, \me{} and \aurora{}, both following a traditional QD loop of selection, variation, evaluation and addition. In this section, we detail the search space, the variation operator, as well as additional design choices to direct the search toward self-organizing, autonomous patterns.

\subsection{Search Space}
We restrict the world configuration to a $2$-dimensional $128 \times 128$ array with $3$ channels and we use a total of $15$ kernels: $3$ self-interacting kernels for each channel and $1$ cross-channel kernel for each pair of distinct channels.
The search space encompasses all possible \emph{genotypes}, which are composed of two elements: the \emph{seed} and the \emph{rule parameters}.

The seed represents the initial configuration of a $32 \times 32 \times 3$ array, totaling $3072$ sites. This array sets the initial configuration for the simulation.
The rule parameters in the genotype are limited to $(\mu_k, \sigma_k, h_k)$ for each kernel, totaling $15 \times 3 = 45$ parameters. The remaining parameters $(r_k R, \beta_k)$ are fixed and shared across all individuals. These parameters can either be randomly sampled at the beginning or taken from an existing self-organizing pattern. In this work, we use the parameters from the soliton ``Aquarium'' (pattern id 5N7KKM), previously discovered by~\citet{chan_LeniaExpandedUniverse_2020}. The initial population consists of a batch of Aquarium solitons with isotropic variations.

The search space spans a total of $3072 + 45 = 3117$ dimensions. The genes of an individual are expressed by initializing a world configuration $\world^0$ to a zero array except for the center, which is initialized with the seed from the genotype. Then, the world is updated according to the genotype's rule parameters for $N$ steps, resulting in a sequence $(\world^0, \dots, \world^{N\Delta t})$, which represents the development and manifestation of the phenotype over time.

\subsection{Variation Operator}
Our methods are genetic algorithms that employ a specialized variation operator known as iso+LineDD~\citep{hypervolume}. This operator is designed to efficiently navigate the genotype space by introducing variations that are informed by both isotropic and directional perturbations. The isotropic variation introduces small, random changes to a solution, ensuring a thorough exploration of the search space around the current solution. In contrast, the directional variation generates new candidate solutions by exploring along the line defined by two existing solutions in the population. Given two parent solutions with gentoypes $\solution_1$ and $\solution_2$, an offspring solution $\solution$ is generated as follows:
$$\solution = \solution_1 + \sigma_1 \mathcal{N}(\mathbf{0}, \mathbf{I}) + \sigma_2 (\solution_2 - \solution_1) \mathcal{N}(0, 1)$$

\subsection{Solitons}
For a given individual, we can decode its genotype into a seed and rule parameters. Its genes are expressed through Lenia simulation: the seed undergoes a developmental process driven by the rule parameters. This developmental process can potentially culminate in the emergence of a distinct, autonomous, self-organizing phenotype, known as a \emph{soliton}. Our focus is on discovering and identifying such solitons that maintain their structure and coherence over time, in contrast with transient, ephemeral or spatially diffused patterns. In this section, we will delve into statistical measures that aid in quantifying the characteristics of patterns and potentially assess their stability and agency.

\subsubsection{Statistical Measures}
Each genotype results in a sequence of world states, $(\world^0, \dots, \world^{N\Delta t})$, from which we can derive various \emph{statistical measures}.
\begin{table}[h]
\centering
\begin{tabular}{ll}
    \toprule
    Statistical measure & Formula\\
    \midrule
    Mass & $m = \sum_{\site \in \lattice} \sum_i \world_i(\site)$\\
    Center of mass & $\overline{\site} = \frac{1}{m} \sum_{\site \in \lattice} \site \sum_i \world_i(\site)$\\
    Velocity & $v = \Delta \overline{\site} / \Delta t$\\
    Angle & $\alpha = \arg(v)$\\
    Linear velocity & $V = |v|$\\
    Angular velocity & $\omega = \Delta \alpha / \Delta t$\\
    Color & $C = \frac{1}{|\lattice|} \sum_{\site \in \lattice} \world(\site)$\\
    \bottomrule
\end{tabular}
\caption{Statistical measures calculated over world state $\world$.}
\label{tab:statistical-measures}
\end{table}
We draw inspiration from the original Lenia paper~\citep{chan2019lenia} where statistical measures are introduced to provide quantitative analyses of phenotypes. We rely on these predefined measures to design fitness and descriptor functions that are manually directed toward characteristics of interest. At each timestep, the world state $\world$ can be transformed into statistical measures, such as mass, velocity, color, see~\Cref{tab:statistical-measures}. For example, the statistical measure ``color'' is a 3-dimensional vector with values ranging from 0 to 1, representing the average RGB (Red, Green, Blue) value of the phenotype. This measure not only conveys the predominant color of the phenotype but also indirectly captures information about its mass. Specifically, an RGB value of $[1, 1, 1]$ indicates white and a maximum mass, while an RGB value of $[0, 0, 0]$ indicates black and an absence of mass.

Furthermore, from the sequence $(\world^0, \dots, \world^{N\Delta t})$, we can derive time-series of statistical measures, denoted $(\mathbf{a}_0, \dots, \mathbf{a}_N)$. In this research, the first $n$ timesteps are considered to be the developmental phase of the seed. After $n$ timesteps, the pattern is presumed stable and the sequence $(\mathbf{a}_{n+1}, \dots, \mathbf{a}_N)$ is used to characterize the phenotype. In particular, this sequence can be aggregated into summary statistics such as mean, median, variance and so on. For instance, velocity sequence $(v_{n+1}, \dots, v_N)$ can be aggregated into a velocity average to assess the typical movement speed or into a velocity variance to quantify the movement stability. These summary statistics facilitate the design of fitness and descriptor functions used in \me{} or \aurora{}. Combined with expert knowledge, these manually defined statistical measures enable to search for solitons with specific characteristics, such as a certain mass, speed or color.

\subsubsection{Unsupervised Statistical Measures}
\begin{figure}
\centering
\includegraphics[width=0.45\textwidth]{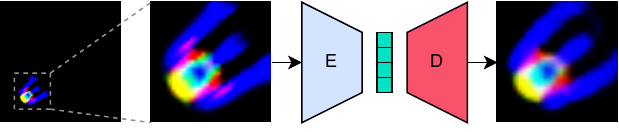}
\caption{The world configuration $\world$ is cropped around the center of mass of the phenotype to form the $32 \times 32 \times 3$ input to the encoder (blue). Then, the encoder compresses the high-dimensional input into a low-dimensional latent vector $\mathbf{z}$ (green). During training, the decoder (red) transforms the latent vector back to the original input to compute a reconstruction loss, that is optimized via gradient descent.}\label{fig:vae}
\end{figure}
Unsupervised representation learning techniques, such as autoencoders, enable to automatically discover statistical measures without requiring labeled data. In our approach, we use a Variational AutoEncoder (VAE)~\citep{vae} to compress high-dimensional phenotypes into lower-dimensional latent representations. These representations are essentially statistical measures that can be used to define fitnesses and descriptors. The VAE architecture comprises two main components: an encoder and a decoder that facilitate the learning of latent representations. The input to the VAE is a $32 \times 32 \times 3$ crop of the configuration $\world$, centered around the center of mass of the phenotype. The encoder transforms the high-dimensional input into a 8-dimensional latent vector $\latent$ and the decoder recreates the input from the encoded representation, see~\Cref{fig:vae}. \aurora{} utilizes data generated during the evaluation step of the QD loop to train the VAE.

Each individual induces a sequence of configurations $(\world^{(n+1)\Delta t}, \dots, \world^{N\Delta t})$ that can be encoded into a latent space trajectory $(\latent_{n+1}, \dots, \latent_N)$ with the encoder. This trajectory within the latent space can be considered as a sequence of statistical measures and can be aggregated into descriptors and fitnesses as outlined in the previous section. The average latent representation provides a succinct and informative summary of a phenotype's essential characteristics. For this reason, we define the unsupervised descriptor as the mean vector of the latent trajectory, $D_\params(\solution) = \frac{1}{N-n} \sum_{i=n+1}^N \latent_i$. This descriptor function can be combined with any manually defined fitness functions.

\begin{figure}
\centering
\includegraphics[width=0.45\textwidth]{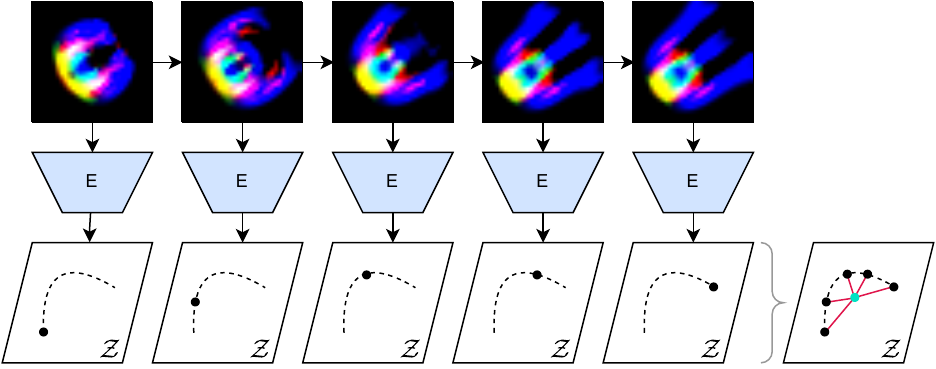}
\caption{The phenotype at different timesteps forms a trajectory in the latent space $\mathcal{Z}$. The green dot represents the mean vector of the latent trajectory, i.e., the unsupervised descriptor of the individual. The red segments represent the Euclidean distance between the latent vectors and the descriptor, used to compute the unsupervised fitness of the individual.}\label{fig:unsupervised-fitness-descriptor}
\end{figure}

Furthermore, we utilize the latent trajectory to develop an unsupervised fitness function that captures fundamental characteristics of life such as agency, self-organization and stability. A key premise is that a stable pattern should manifest minimal variance in its latent representation. A latent representation provides a condensed yet informative abstraction of the raw pixel data. Consequently, variance in the latent representation reflects meaningful changes in the phenotype's structure, rather than superficial image noise or minor visual discrepancies. To quantify this, we define the unsupervised fitness as the negative average Euclidean distance between the latent vectors and the mean vector of the trajectory, $F_\params(\solution) = -\frac{1}{N-n} \sum_{i=n+1}^N ||\latent_i - D_\params(\solution)||_2$, see~\Cref{fig:unsupervised-fitness-descriptor}. This fitness measures the spread or dispersion of the latent vectors relative to their mean in a multidimensional space. This approach is grounded in the principle that ``what persists, exists'' and is related to homeostasis, one of the pillars of life~\citep{lyfe}.

These unsupervised descriptor and fitness functions are particularly valuable because their computation within a latent space makes them domain-agnostic. However, the fitness function is only partially unsupervised as it is intentionally biased to capture the spread of the latent trajectory. Moreover, this definition of self-organization and stability has limitations, as it tends to penalize individuals exhibiting periodic stability or engaging in chaotic movements.

\subsubsection{Constraints on Growth}
To direct the search toward stable patterns, we enforce three constraints on the phenotypes. The objective of these constraints is to ensure that phenotypes do not exhibit positive or negative infinite growth~\citep{chan2019lenia}. Explosion or evaporation happens when the mass expands to very large values or shrinks to zero. We discard any individuals inducing a sequence $(\world^{(n+1)\Delta t}, \dots, \world^{N\Delta t})$, where at least one phenotype has evaporated, exploded or is too spread. These constraints are controlled through three hyperparameters: A minimum and maximum mass threshold $m_{\text{min}}$, $m_{\text{max}}$ and a mass spread $\sigma_m$.

\section{Experiments}
The objective of our experiments is threefold. First, we assess the capability of \ours{} to evolve a population of individuals with a diversity manually directed toward specific characteristics of interest. Second, we evaluate the ability of \ours{} to evolve an unsupervised diversity of individuals, illuminating Lenia's vast landscape. Third, we explore \ours{}'s potential to exhibit characteristics of open-ended evolution, such as unbounded diversity.

We conducted a series of experiments using the QDax framework and a GPU-accelerated implementation of Lenia in JAX. Each experiment is replicated 10 times with random seeds. We report $p$-values based on the Wilcoxon–Mann–Whitney $U$ test with Holm-Bonferroni correction. The source code is available at \href{https://leniabreeder.github.io}{leniabreeder.github.io}. Population size is $32,768$ for all experiments.

\subsection{Manual Diversity}
\begin{figure*}
    \centering
    \includegraphics[width=0.92\textwidth]{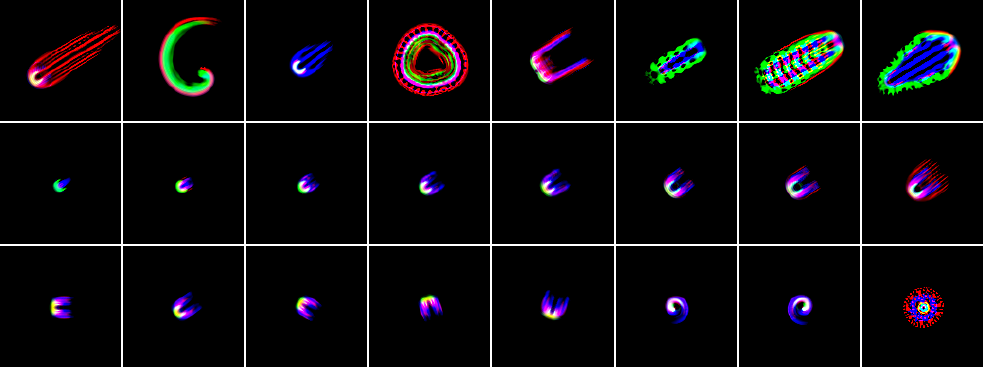}
    \caption{\textbf{\me{}} Each row displays a single, independent run with each image sized $128 \times 128 \times 3$. \textbf{Row 1} features individuals selected for \emph{velocity average} fitness and \emph{color} descriptor, arrayed from left to right to showcase proximity to specific colors such as red $[1, 0, 0]$, green $[0, 1, 0]$, blue $[0, 0, 1]$, red-green, red-blue, blue-green, red-green-blue and random $[0.01, 0.6, 0.5]$. \textbf{Row 2} focuses on \emph{negative angle variance} for fitness with \emph{mass} and \emph{velocity} as descriptors, showing a gradient of increasing mass and constant velocity. \textbf{Row 3} highlights \emph{negative mass variance} for fitness with \emph{angle} and \emph{velocity} as descriptors, arranging samples by different angles, clockwise rotation, counterclockwise rotation and no rotation.}
    \label{fig:me}
\end{figure*}
We report five experiments using \me{}, each employing different combinations of manually defined descriptors and fitnesses, summarized in~\Cref{tab:me-results}.
This algorithm necessitates choosing appropriate bounds for each statistical measure. For color and angle, bounds naturally encompass all potential values, with color in the range $[0, 1]^3$ and angle in $[-\pi, \pi]$. However, for mass and velocity, we choose arbitrary yet reasonable bounds of $[0, 16]$ and $[0, 0.5]$ respectively, due to the absence of natural limits. For context, the soliton named Aquarium (pattern id 5N7KKM), discovered by~\citet{chan_LeniaExpandedUniverse_2020}, has a mass of roughly 2.42 units and a velocity of 0.12 units.
To evaluate \me{}, we consider two metrics. The \emph{coverage} corresponds to the proportion of filled cells in the grid of solutions and the \emph{max fitness} corresponds to the fitness of the most optimal solution in the grid.

\begin{table}
\centering
\tabcolsep=0.1cm
\begin{tabular}{llccc}
    \toprule
    Fitness & Descriptor & Coverage & Max fitness\\
    \midrule
    velocity avg & [color] & $38.5\%$ & $1.01$\\
    mass avg & [color] & $39.8\%$ & $97.5$\\
    mass var & [color] & $39.0\%$ & $1,525$\\
    neg angle var & [mass, velocity] & $52.2\%$ & $-1.31 \times 10^{-7}$\\
    neg mass var & [angle, velocity] & $42.7\%$ & $-2.00 \times 10^{-7}$\\
    \bottomrule
\end{tabular}
\caption{\textbf{\me{}} Median coverage and max fitness for different combinations of fitness and descriptor functions.}
\label{tab:me-results}
\end{table}

The experimental results reported in~\Cref{tab:me-results} indicate that \me{} effectively generates diverse individuals with targeted characteristics. For example, \me{} successfully identified solitons across a broad spectrum of colors combined with three distinct fitness functions. Specifically, the coverage of approximately $39\%$ of the color space is particularly significant considering that not all possible colors are viable. For instance, pure black $[0, 0, 0]$, indicating patterns that have evaporated, and pure white $[1, 1, 1]$, indicating exploded patterns, are inherently unattainable. The qualitative evidence in~\Cref{fig:me} confirms \me{}'s capability to discover patterns spanning all intended color combinations.

Moreover, \me{} efficiently identified solitons with desired locomotion traits as defined by various combinations of fitness and descriptor functions. The experiments demonstrated effective optimization of fitnesses such as negative angle and mass variances, as indicated by max fitness values approaching zero. This method is able to evolve a population of species with a wide range of masses, different velocities and different orientations as indicated by the high coverage score. The qualitative results in~\Cref{fig:me} demonstrate that this approach is able to find patterns with a wide range of mass, as well as different locomotion angles and angular velocities.

\subsection{Unsupervised Diversity}
\begin{figure*}[!ht]
    \centering
    \includegraphics[width=0.92\textwidth]{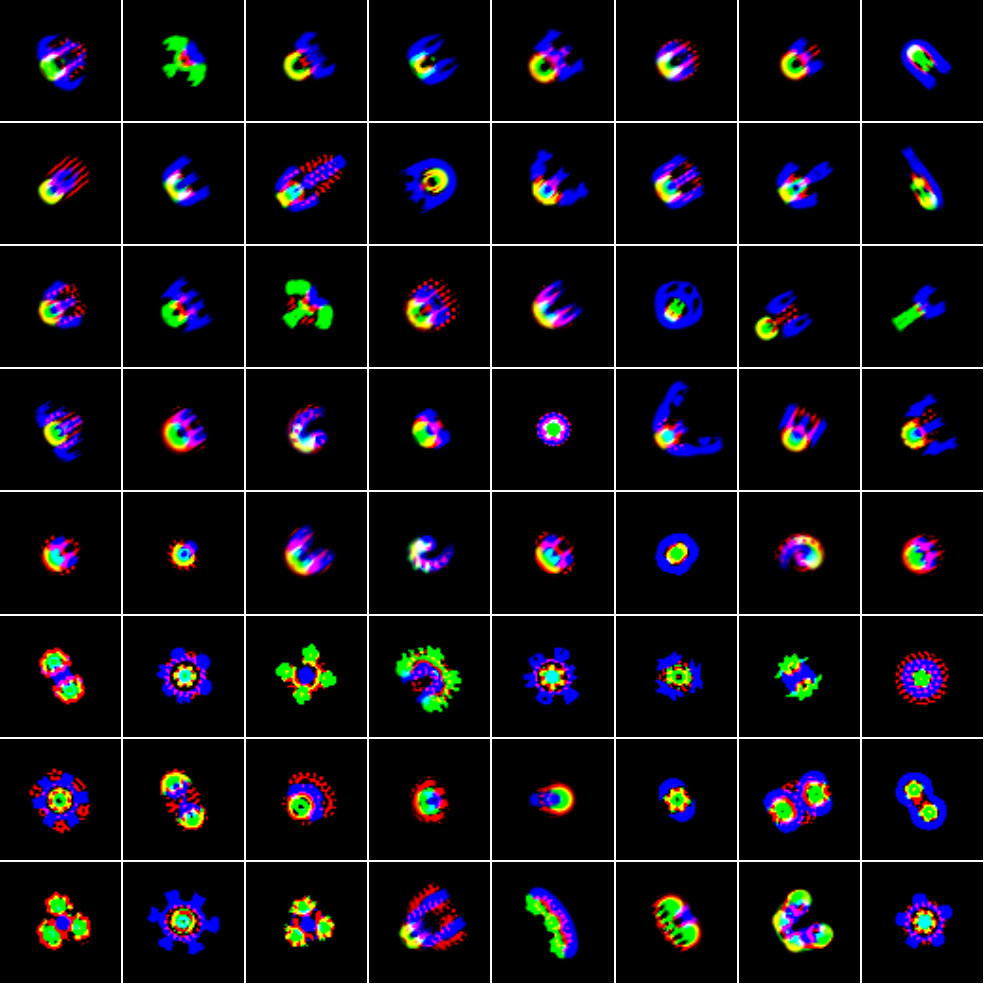}
    \caption{\textbf{\aurora{}} Each block of rows displays a single, independent run with each image sized $64 \times 64 \times 3$. \textbf{Row 1-3} Fitness is the \emph{negative angle variance}. \textbf{Row 4-5} Fitness is the \emph{negative mass variance}. \textbf{Row 6-8} Fitness is \emph{unsupervised}.}
    \label{fig:aurora}
\end{figure*}
\begin{figure*}[h!]
    \centering
    \includegraphics[width=0.92\textwidth]{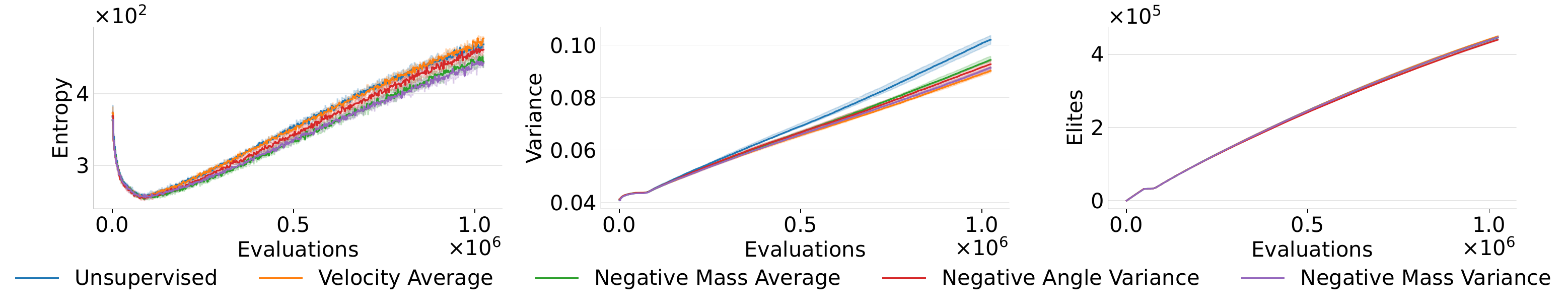}
    \caption{\textbf{\aurora{}} Entropy, variance and cumulative elites with different fitness functions. The solid line is the median and the shaded area represents the first and third quartiles.}
    \label{fig:plot_aurora}
\end{figure*}
We report five experiments using \aurora{}, pairing the unsupervised descriptor with various fitness functions including velocity, negative mass, negative angle variance, negative mass variance, and a purely unsupervised approach.
To evaluate \aurora{}, we consider three metrics. The \emph{entropy} corresponds to an estimation of the amount of information in the population and is related to the notion of ecology, introduced by~\citet{dolson_MODESToolboxMeasurements_2019} to quantify diversity. We utilize the VAE that models the likelihood and posterior distributions of the data $X$ (i.e., the phenotypes), to develop a Monte-Carlo approximation of the entropy via the formula $H(X) = H(Z) + H(X | Z) - H(Z | X)$. Note that the accuracy of this metric depends on the quality of the VAE.
The \emph{variance} metric measures the average pixel-wise variance among the phenotypes stored in the repertoire, serving as a coarse but direct indicator of population diversity. In contrast with entropy, this practical metric provides a robust estimation that is independent of the VAE, making it a reliable and grounded measure of diversity.
Finally, the \emph{cumulative elites} tracks the cumulative sum of novel offspring solutions added to the repertoire during the run, reflecting the influx of novel solitons that continually refresh the population.

The qualitative results in~\Cref{fig:aurora} illustrate that \aurora{} can evolve an expansive diversity of solitons, encompassing a wide array of shapes, colors, sizes, and locomotion properties --- significantly broader than the targeted diversity seen in \me{}.
In \Cref{fig:plot_aurora}, initial entropy values stabilize as the autoencoder, which starts with random weights, begins to train. Subsequently, entropy consistently increases, indicating a growing information richness within the population, accompanied by an increase in phenotype variance. This suggests that the repertoire is continually enriched with novel phenotypes exhibiting different shapes and colors.
Notably, runs employing unsupervised fitness consistently show higher variance (statistically significant with $p < 0.005$), suggesting a more diverse array of traits compared to those with predefined fitness functions.
The steady increase in the cumulative elites added to the population, even after one million evaluations, indicates a growing number of novel individuals constantly being found, demonstrating a continuous change in the information content of the population, which supports the sustained generation of novelty~\citep{dolson_MODESToolboxMeasurements_2019}. This constant flow shows that the dynamic interaction between the population and the niche boundaries fostered by \aurora{} is effective in promoting a diverging and ever-evolving ecosystem, that avoids premature convergence.
The ongoing increase in population entropy and variance, coupled with the continuous introduction of new elites, highlights \aurora{}'s potential to drive open-ended evolution, aligning with some of the key dynamics --- namely, the perpetual production of novelty, unbounded diversity, and continuous change in information content~\citep{dolson_MODESToolboxMeasurements_2019,packard_OverviewOpenEndedEvolution_2019,soros_IdentifyingNecessaryConditions_2014}.

\section{Conclusion}
We show that Quality-Diversity is an effective framework for the automatic discovery of \emph{diverse} self-organizing patterns in complex systems.
Our findings not only showcase the breadth of artificial life within Lenia but also underscore the relevance of Quality-Diversity algorithms in illuminating an ecosystem of artificial species and exhibiting a sustained generation of diversity.
Combined with Lenia, we show that Quality-Diversity has the potential to present some hallmarks of open-ended evolution, aligning with its original purpose.
We make the code publicly available and encourage the community to participate in the discovery of novel and diverse life forms.

At the core of \ours{} lies the utilization of a novel unsupervised fitness function. Yet, it relies on simple heuristics that only mimics homeostasis. We posit that enhancing this fitness function would enable to discover even more meaningful expressions of artificial life.
Furthermore, the current autoencoder architecture is not invariant to rotation or scaling. We believe that improving the autoencoder architecture could also benefit the framework to capture a more refined notion of diversity.
A direction for future work is to quantify the impact of different starting patterns and explore whether diverse evolutionary trees can be discovered by beginning with alternative solitons.

\footnotesize
\bibliographystyle{apalike}
\bibliography{bibliography} 


\end{document}